\pdfoutput=1
\documentclass{article}
\usepackage[dblblindworkshop, final]{neurips_2026}
\usepackage[utf8]{inputenc}
\usepackage[T1]{fontenc}
\usepackage{hyperref}
\usepackage{url}
\usepackage{booktabs}
\usepackage{amsfonts}
\usepackage{amsmath}
\usepackage{nicefrac}
\usepackage{microtype}
\usepackage{xcolor}
\usepackage{graphicx}
\usepackage{multirow}

\newcommand{\corpus}{the pooled set}

\title{A Function-Level Vulnerability Score Measures Flag Rate More Than the Model: Protocol Effects on Paired Benchmarks}

\workshoptitle{TAE (Trust-AI-Eval): Can We Trust AI Evaluation?}

\author{%
  \begin{tabular}{c@{\hspace{3em}}c}
  Maciej Cichoń & Bartłomiej Dmitruk\\[2pt]
  \mdseries University of Lodz & \mdseries Warsaw University of Technology\\
  \mdseries Striga & \mdseries Striga\\[3pt]
  \multicolumn{2}{c}{\mdseries\texttt{\{maciej,bartek\}@striga.ai}}
  \end{tabular}
}

\begin{document}

\maketitle

\begin{abstract}
Language models are increasingly evaluated as vulnerability detectors, and scores reported for similar models differ widely between papers. We measured how much of that difference evaluation protocol accounts for, with model outputs held fixed. In a paired test, a model must flag a vulnerable function and clear its version after a fixing commit. Three choices that published evaluations make differently were varied one at a time: metric, verdict extraction and output budget. Seven frontier and large open models were evaluated on five released pair benchmarks and a set pooled for this work under one protocol, and 61 open models of 1.5B to 36B parameters on the pooled set. Function-level F1 follows how often a model flags both functions of a pair (Spearman $+0.86$ over 42 combinations) and is nearly unrelated to pair-level correctness ($+0.16$). On the pair score, extraction changes a model's number by $+0.001$ at the median and budget by $+0.02$ with an interval through zero, whereas the model changes a benchmark's number by up to 0.18 and the benchmark a model's by up to 0.16; a function-level score therefore measures flag rate more than model. For 37 of 68 models the difference between correct and reversed pairs is within its 95\% interval of zero, the value for a null model that flags each function at a fixed rate, while both-flagged and both-cleared rates exceed that null by 0.055 on median, and for 64 of 68 both functions of a pair receive one answer more often than independence predicts: verdicts are determined by the text common to both functions. On length-matched pairs a linear probe on activations separates 0.78 by within-pair ranking, against 0.64 for a tf-idf baseline and 0.5 for length; the generated verdict is near chance for three of six models and at 0.55 to 0.57 for the other three, and a prompted logit is at chance for all six.
\end{abstract}

\section{Introduction}

Most reported results on language models for vulnerability detection consist of one number per model, accuracy or F1 over functions labelled vulnerable or benign. PrimeVul \citep{ding2024primevul} introduced the paired version of the test. Each vulnerable function is paired with its version after the fixing commit; these two functions are the sides of a pair. A model is credited only when it flags the first and clears the second. Several benchmarks have since adopted pairs, each with a protocol of its own.

What does the reported number depend on once the model is fixed? Three choices differ between published evaluations. The present work varies each of them while model outputs stay the same:
\begin{enumerate}
\item the metric: function-level accuracy or F1, or the four outcomes of a pair;
\item verdict extraction: a deterministic parse of a required verdict line, or a language-model judge that is given the analysis, as in SecLLMHolmes \citep{ullah2024secllmholmes};
\item output budget: PrimeVul's released configuration caps the chain of thought at 1,024 tokens, whereas current work uses 16k to 32k \citep{openvul2026,li2025vulpo}.
\end{enumerate}
Two further measurements use the same outputs. The released pair benchmarks (PrimeVul, VulnPatchPairs, PairVul, JitVul, SecLLMHolmes and a set pooled for this work) were evaluated under one protocol with seven API models. Sixty-one self-hosted open models of 1.5B to 36B parameters were evaluated on the pooled set. For six open models the same pairs were also scored by a prompted yes/no logit, a released activation oracle and a supervised linear probe on activations.

Three statements follow, developed in Section~\ref{sec:disc}. Verdicts are determined by the text common to both sides. The flag rate, whose contribution $p(1-p)$ does not depend on size, removes most of the difference between models of very different size. A linear probe on the activations separates three quarters of pairs by ranking where the generated verdict does not, and the signal it uses follows code that a patch adds. The low scores are therefore not explained by the function text carrying no usable information. A function-level score consequently measures the flag rate more than the model, whereas on the pair score the model and the benchmark matter more than extraction or budget (Table~\ref{tab:effects}).

\section{Related work}

Function-level vulnerability detection has been evaluated on datasets built from security-fix commits, and the problems of those datasets are documented. \citet{croft2023dataquality} found 20 to 71 percent of labels inaccurate and widespread duplication, \citet{chakraborty2022reveal} and \citet{steenhoek2023empirical} showed that classifiers exploit regularities of the data, and \citet{risse2025topscore} found on manual inspection that many functions are vulnerable only through their calling context. PrimeVul \citep{ding2024primevul} introduced the paired test, and VulnPatchPairs \citep{risse2024limits}, PairVul \citep{du2024vulrag}, JitVul \citep{jitvul2025}, SecLLMHolmes \citep{ullah2024secllmholmes} and OpenVul \citep{openvul2026} followed with pairs of their own. In the pair-level results reported for JitVul and PairVul, frontier models lie within four to six points of 8B and 32B open models, and on CWE-Trace every model from 4B to GPT-4.1-mini scores 49 to 54 percent (Appendix~\ref{app:lit}); \citet{zibaeirad2026directional} traced the low scores to models giving one answer for both functions of a pair. The present work measures that tendency against a fixed-rate null model.

That a protocol choice can produce an effect absent from the model outputs has a precedent: \citet{schaeffer2023emergent} showed that emergent abilities appear and disappear with the metric. On extraction, \citet{yu2024xfinder} showed that regular expressions miss answers a trained extractor finds, and \citet{tan2024judgebench} that language-model judges are near chance on hard items.

Linear probes \citep{alain2016probes} report what a classifier can decode from activations, which is not the same as what the model uses \citep{hewitt2019control,belinkov2022probing,ravichander2021probing,elazar2021amnesic}. For truthfulness, decodable directions exist \citep{burns2023latent,marks2024geometry} and change the output when intervened on \citep{li2023iti,marks2024geometry}, and models have been shown to encode more than they say \citep{azaria2023internal,orgad2025know}. The probe result here is of that form.

\section{Setup}
\label{sec:setup}

\paragraph{Pair benchmarks.} A pair is a vulnerable function and its version after the fixing commit, its two sides. PrimeVul's paired split, VulnPatchPairs, PairVul and JitVul form pairs from security-fix commits with different filters. SecLLMHolmes uses hand-curated real and synthetic pairs. The set pooled for this work holds 5,705 C/C++ pairs from three sources: 1,756 pairs from the paired data of PrimeVul across its splits \citep{ding2024primevul}, 1,408 pairs cut from the reproduced OSS-Fuzz vulnerabilities of ARVO \citep{mei2024arvo} and their located fixes, and 2,541 pairs extracted by the authors from CVE-fixing commits. Its execution-verified tier, 1,541 pairs, consists of the ARVO pairs, taken as published, and 133 of the commit pairs, which the authors reproduced under a sanitizer. The other 4,164 pairs carry commit-heuristic labels like the other benchmarks. Pair construction and counts after exact-duplicate removal are given in Table~\ref{tab:benches} (Appendix~\ref{app:benches}).

\paragraph{Protocol.} Every model received the same system prompt and user template (Appendix~\ref{app:prompt}). The prompt requests a first line that is exactly \texttt{Verdict: vulnerable} or \texttt{Verdict: not vulnerable}, followed by a short justification. Each side of each pair was sampled $K=3$ times at temperature 0.7. The majority verdict was taken. Ties and missing verdict lines were recorded as separate outcomes. The output budget was 32,768 tokens. For the seven API models, this budget gave the same result as no cap on a 40-pair sample. Each released benchmark was drawn by a seeded shuffle and capped at 200 pairs, all 33 for SecLLMHolmes. The pooled set was drawn per tier: all 133 pairs reproduced by the authors were taken, and each tier was filled to 500 by a seeded round-robin over projects, which favours sources with many projects. The draw therefore holds 367 ARVO pairs and the 133 author pairs in its verified half and 423 PrimeVul pairs and 77 commit pairs in its heuristic half; 23 of its PrimeVul pairs are also in the 200-pair PrimeVul draw. No model was trained or tuned; the probes of Section~\ref{sec:methods} were fitted on activations.

\paragraph{Scores.} The two verdicts of a pair fall into one of four outcomes: P-C, vulnerable side flagged and patched side cleared; P-V, both flagged; P-B, both cleared; P-R, reversed. Pair-correct is the P-C rate, reported with a bootstrap 95\% interval. The \emph{net score} is P-C minus P-R. Function-level accuracy and F1 treat the $2n$ sides as independent items. The reference for these scores is the null model that flags each side with probability $p$, whichever side it is shown. It produces
\begin{equation}
\mathrm{P\text{-}V} = p^2, \qquad \mathrm{P\text{-}B} = (1-p)^2, \qquad \mathrm{P\text{-}C} = \mathrm{P\text{-}R} = p(1-p).
\label{eq:null}
\end{equation}
Its net score is zero for every $p$. Its pair-correct is 0.25 at $p = 0.5$. On a balanced set of sides its F1 is $2p/(1+2p)$, which is 0.67 when it flags everything.

\paragraph{Models.} Seven models were accessed through API: claude-opus-4.8, gpt-5.2, grok-4.3, deepseek-r1, deepseek-v3.2, qwen3-coder and qwen3-8b. Two open models were additionally self-hosted over all 5,705 pairs, qwen3-8b in BF16 and DeepSeek-V4-Flash in its released FP8/FP4 checkpoint \citep{kurtic2024fp8}. Sixty-one open models of 1.5B to 36B parameters from thirteen base families were self-hosted on the 1,000-pair draw. They comprise general-purpose instruction models, code-tuned and security-tuned checkpoints, reasoning distillations, community fine-tunes, abliterated variants of eight models, from which the refusal direction was removed, and one base checkpoint. Where a model's context was shorter than 36,864 tokens, its budget was the context minus 4,096. Six of these models (Qwen3 1.7B, 4B, 8B and 14B, Llama-3.1-8B-Instruct, gemma-2-9b-it) have a released activation oracle \citep{karvonen2025oracles}. The full list, with family, budget and unparsed rate, is Table~\ref{tab:population} (Appendix~\ref{app:pop}).

\section{Results}
\label{sec:results}

Table~\ref{tab:effects} collects the effect of every choice in pair-correct units; the subsections give the measurements behind each row.

\begin{table}[t]
\centering
\caption{Effect of each choice on pair-correct, with the effect of the model and of the benchmark for comparison. Intervals are 95\% bootstrap intervals over pairs; ranges are over models or benchmarks.}
\label{tab:effects}
\resizebox{\textwidth}{!}{\begin{tabular}{llll}
\toprule
Choice & Comparison & Change in pair-correct & Basis \\
\midrule
Metric & function-level F1 in place of pair-correct & $+0.29$ to $+0.59$ (median $+0.44$) & 42 combinations \\
Extraction & language-model judge in place of the parse & median $+0.001$, range $-0.11$ to $+0.17$ & 68 models, three judges \\
Budget & 32,768 tokens in place of 1,024, gpt-5.2 & $+0.022$ ($-0.022$ to $+0.067$) & 90 pairs \\
Serving stack & API in place of self-hosted, qwen3-8b & $+0.008$ ($-0.021$ to $+0.038$) & 1,000 pairs \\
Model & range over seven models, one benchmark & 0.07 to 0.18 & six benchmarks \\
Model & range over 61 open models & 0.00 to 0.27 & pooled set \\
Benchmark & range over six benchmarks, one model & 0.05 to 0.16 & seven models \\
\bottomrule
\end{tabular}}
\end{table}

\subsection{The metric}
\label{sec:metric}

\begin{figure}[t]
\centering
\includegraphics{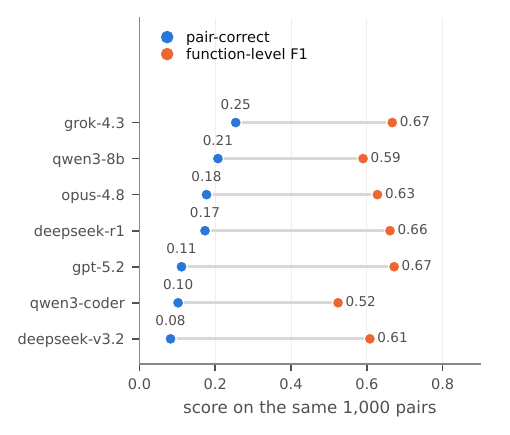}
\caption{F1 and pair-correct order the same outputs differently. Pair-correct and function-level F1 for the seven API models on the 1,000-pair draw of \corpus{}, sorted by pair-correct.}
\label{fig:metric}
\end{figure}

The first of the three choices is the metric. Pair-correct lies between 0.06 and 0.25 for every combination of benchmark and model. F1 on the same outputs lies between 0.36 and 0.68 (Figure~\ref{fig:metric}; Tables~\ref{tab:cells} and~\ref{tab:panel}). Over the 42 combinations, the Spearman correlation between F1 and the both-flagged rate P-V is $+0.86$, with a bootstrap interval over combinations of $+0.74$ to $+0.92$. Between F1 and pair-correct it is $+0.16$, with an interval of $-0.16$ to $+0.46$. Within a benchmark, the two orders of the models correlate between $-0.14$ on JitVul and $+0.46$ on PrimeVul. gpt-5.2 obtained the highest or joint highest F1 on five of the six benchmarks, with values of 0.56 to 0.67. Its pair-correct on the same five is 0.08 to 0.12. qwen3-coder cleared both sides of 45\% to 73\% of pairs. It obtained the lowest F1 of the panel and a pair-correct comparable to that of gpt-5.2, above it on three benchmarks and below on three. F1 follows flag rate, as Equation~\ref{eq:null} predicts.

\begin{table}[t]
\centering
\caption{The four pair outcomes, the share of pairs unparsed or tied, and function-level F1 on identical outputs, pooled-set draw ($n$=1,000 pairs; the two self-hosted rows cover all 5,705). P-C: pair-correct; P-V: both sides flagged; P-B: both sides cleared; P-R: reversed; net score: P-C minus P-R.}
\label{tab:cells}
\resizebox{\textwidth}{!}{
\begin{tabular}{lccccccc}
\toprule
Model & P-C & P-V & P-B & P-R & net score & unparsed or tied & function-level F1 \\
\midrule
opus-4.8 & 0.177 & 0.533 & 0.271 & 0.019 & 0.158 & 0.000 & 0.628 \\
gpt-5.2 & 0.111 & 0.819 & 0.051 & 0.019 & 0.092 & 0.000 & 0.672 \\
grok-4.3 & 0.254 & 0.530 & 0.179 & 0.037 & 0.217 & 0.000 & 0.667 \\
deepseek-r1 & 0.173 & 0.670 & 0.100 & 0.047 & 0.126 & 0.010 & 0.661 \\
deepseek-v3.2 & 0.082 & 0.674 & 0.188 & 0.056 & 0.026 & 0.000 & 0.608 \\
qwen3-coder & 0.102 & 0.411 & 0.453 & 0.034 & 0.068 & 0.000 & 0.524 \\
qwen3-8b & 0.207 & 0.443 & 0.237 & 0.112 & 0.095 & 0.001 & 0.590 \\
\midrule
qwen3-8b, self-hosted, all 5,705 pairs & 0.189 & 0.418 & 0.294 & 0.098 & 0.091 & 0.001 & 0.572 \\
deepseek-v4-flash, self-hosted, all 5,705 pairs & 0.084 & 0.823 & 0.021 & 0.010 & 0.074 & 0.063 & 0.683 \\
\bottomrule
\end{tabular}
}
\end{table}

The four outcomes separate two failure modes that one number merges. The reversed rate adds a second piece of information. On \corpus{}, qwen3-8b has P-C 0.207 and P-R 0.112. opus-4.8 has 0.177 and 0.019. By pair-correct the 8B model ranks above opus on four of the six benchmarks and ties on one; by net score it is above on two, ties on one and is below on three, because its reversed rate is 0.06 to 0.12 against 0.00 to 0.04 for opus. Net score is the input-dependent part of pair-correct.

The null model of Equation~\ref{eq:null} flags each side with a fixed probability $p$ and answers both sides independently. A coin, $p = 0.5$, has a pair-correct of 0.25, above 65 of 68 models, so the coin is not the right null. The null's net score is zero, and for 37 of 68 models the 95\% interval of the observed net score includes zero (Table~\ref{tab:null} and Figure~\ref{fig:population}, Appendix~\ref{app:null}); their flag rates span the whole range, from 0.03 to 1.00. Their other outcome rates do not match the null: with $p$ set to each model's flag rate and all rates taken over parsed pairs, the both-flagged rate exceeds $p^2$ by 0.055 on median and the both-cleared rate exceeds $(1-p)^2$ by 0.055, and no model is below either value. Models do not answer both sides independently. For 64 of 68, a Fisher exact test on both verdicts of a pair rejects independence at the 0.05 level, with Cohen's $\kappa$ between 0.08 and 0.74; three of the remaining four flag every side. Verdicts therefore depend on text both sides share. The patch, the only difference between sides, does not change it. The 30 models whose net score is above zero beyond its interval have flag rates from 0.16 to 0.96 and net scores of 0.03 to 0.22; one model, an abliterated Llama, is below zero.

\paragraph{Model strength.} The net score averaged over six benchmarks (Table~\ref{tab:net}) is 0.114 for grok-4.3 and opus-4.8, 0.093 for deepseek-r1, 0.090 for qwen3-8b, 0.056 for gpt-5.2, 0.050 for qwen3-coder and 0.038 for deepseek-v3.2. Within a benchmark the seven models span 0.07 to 0.18 in pair-correct and 0.11 to 0.23 in F1, so neither metric separates the models more than the other; what differs is their order. On the pooled-set draw the strongest open models reach 0.16 to 0.22 (Section~\ref{sec:panel}), equal to grok-4.3 and above opus-4.8, and for 37 of the 61 open models the interval of the net score includes zero.

\subsection{Verdict extraction}
\label{sec:extraction}

The second choice is how a verdict is read from an output. Three open judges from three families (Llama-3.3-70B-Instruct in FP8, gpt-oss-120b, Qwen3-32B) were given every sample produced on \corpus{} by every model, API and self-hosted alike. Each judge received the full analysis (Appendix~\ref{app:judges}). Where a sample contains a verdict line, the judges agreed with it on a median of 99.9\%, 96.8\% and 99.9\% of samples per model. Pair-correct under judge extraction differed from pair-correct under the parse by a median of $+0.001$. The abliterated gpt-oss-20b writes ``vulnerable'' on both sides of 99\% of pairs. Judges agree with its verdict line on 63 to 66\% of samples and raise its pair-correct from 0.006 to 0.16 to 0.18. Qwen3.6 models are strongest in the population when parsed. They lose 0.06 to 0.11 under the gpt-oss judge and 0.00 to 0.03 under the other two. The same judge takes gpt-5.2 from 0.111 to 0.064 and GLM-4.7-Flash from 0.132 to 0.057. Its disagreements run almost entirely one way, a parsed ``not vulnerable'' read as vulnerable, and about two thirds of them fall on the patched side (450 of 728 for Qwen3.6-27B), so they turn correct pairs into both-flagged pairs. On the samples without a verdict line, about 5,200 per judge, judges answered ``vulnerable'' for 44 to 57\% and ``not vulnerable'' for 43 to 55\%, and gave no answer for 0.2 to 0.3\%. Parsing fails on three output formats, where a judge or a tolerant parse changes pair-correct by 0.03 to 0.10 (Appendix~\ref{app:judges}). Extraction changes the number only where text and verdict line disagree or where no verdict line exists. There, the number depends on which judge is used.

\subsection{Output budget}
\label{sec:budget}

The third choice is output budget, which determines how many samples end with a verdict line. The panel was evaluated at 1,024, 4,096 and 32,768 output tokens. At 1,024 tokens, the value of PrimeVul's released configuration, gpt-5.2 produced no verdict line for 19.3\% of its samples over six benchmarks. The share was 0.6\% at 4,096 tokens and zero at 32,768. At 1,024 tokens, 90 pairs of the draw were evaluated; on those, pair-correct was 0.067 at 1,024 tokens against 0.089 at 32,768, a difference of $+0.022$ with a bootstrap interval of $-0.022$ to $+0.067$. deepseek-r1 produced no verdict line for 11.0\% of samples at 4,096 tokens and 0.6\% at 32,768. DeepSeek-V4-Flash reached the 32,768-token limit in 10\% of its samples. The budget at which fewer than 1\% of samples lack a verdict line is therefore 4,096 tokens for gpt-5.2, 32,768 for deepseek-r1 and above 32,768 for DeepSeek-V4-Flash.

\subsection{Benchmarks and the open population}
\label{sec:panel}

\begin{table}[t]
\centering
\caption{Pair-correct per benchmark and model under the protocol (32,768-token budget, $K$=3, deterministic parse). $n$ is the number of drawn pairs; bootstrap intervals are about $\pm 0.05$ at $n$=200 and $\pm 0.025$ at $n$=1,000.}
\label{tab:panel}
\resizebox{\textwidth}{!}{
\begin{tabular}{lccccccc}
\toprule
Benchmark & opus-4.8 & gpt-5.2 & grok-4.3 & deepseek-r1 & deepseek-v3.2 & qwen3-coder & qwen3-8b \\
\midrule
PrimeVul ($n$=200) & 0.09 & 0.08 & 0.18 & 0.17 & 0.07 & 0.07 & 0.17 \\
VulnPatchPairs ($n$=200) & 0.13 & 0.08 & 0.15 & 0.12 & 0.09 & 0.12 & 0.13 \\
PairVul ($n$=200) & 0.19 & 0.12 & 0.18 & 0.18 & 0.15 & 0.06 & 0.14 \\
JitVul ($n$=200) & 0.12 & 0.09 & 0.17 & 0.15 & 0.11 & 0.10 & 0.23 \\
SecLLMHolmes ($n$=33) & 0.15 & 0.06 & 0.09 & 0.12 & 0.12 & 0.09 & 0.24 \\
Ours ($n$=1000) & 0.18 & 0.11 & 0.25 & 0.17 & 0.08 & 0.10 & 0.21 \\
\bottomrule
\end{tabular}
}
\end{table}

Do the six benchmarks agree on the seven models? Kendall's $W$ on the pair-correct order is 0.65. Random orders give 0.33 at the 95th percentile ($p$ = 0.0002).  The pooled set agrees with PrimeVul at 0.93, partly by construction, since 423 of its 1,000 drawn pairs are PrimeVul data and 23 are in the PrimeVul draw itself, and with VulnPatchPairs at 0.81 and JitVul at 0.79. grok-4.3 ranked first on three benchmarks, qwen3-8b on two and opus-4.8 on one. No benchmark placed gpt-5.2 above fifth. Adjacent models in the pair-correct order are separated, by a paired bootstrap over the same pairs, for 3 of the 36 adjacent pairs across the six benchmarks (deepseek-r1 above opus-4.8 on PrimeVul; grok-4.3 above qwen3-8b and deepseek-r1 above gpt-5.2 on the pooled set). The benchmarks agree on the coarse order and separate few neighbours.

\begin{figure}[t]
\centering
\includegraphics{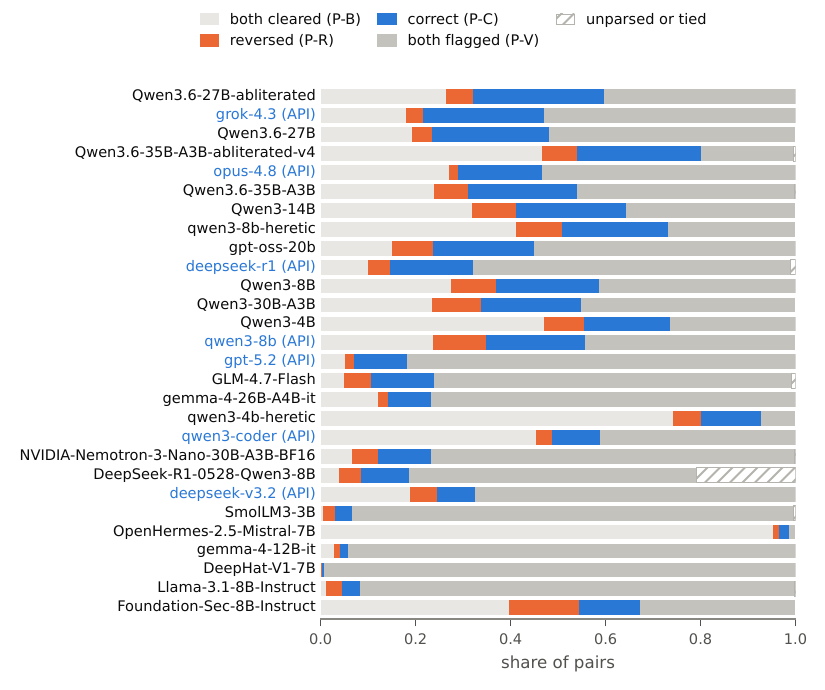}
\caption{The strongest open models match the API models, and most of the population flags or clears nearly every pair. The four pair outcomes on the 1,000-pair draw of \corpus{} for the seven API models (blue names), the fifteen open models with a net score above 0.05, and six open models that flag or clear nearly every pair, sorted by net score; the hatched remainder is pairs unparsed or tied. All 68 models are shown in Figure~\ref{fig:cells} (Appendix~\ref{app:pop}).}
\label{fig:cellsbody}
\end{figure}

The open population extends the panel downwards in size (Figure~\ref{fig:cellsbody}; Table~\ref{tab:popcells}, Appendix~\ref{app:pop}). Pair-correct ranged from 0.00 to 0.27. The net score ranged from $-0.03$ to $+0.22$, and for 37 of the 61 models the interval of the net score includes zero. A quarter of models flagged both sides of more than 80\% of pairs, and a smaller group cleared both sides of most pairs. Fifteen models exceeded a net score of 0.05. Eleven of them are Qwen checkpoints. Security tuning did not raise the number: seven security-tuned checkpoints lie between $-0.03$ and $+0.04$, three of them below zero. The base checkpoint of Qwen3-8B has a net score of $-0.02$. Its instruct version has 0.12. Abliterated variants kept their base's net score within 0.035 in seven of eight cases and changed its default answer in a family-dependent direction (Appendix~\ref{app:twins}); the two best open models are an abliterated Qwen3.6-27B and its base, within noise of each other, so abliteration neither raises nor lowers the score at the top.

\section{Four methods of obtaining a verdict from the same model}
\label{sec:methods}

A low score can belong to a model or to the way its verdict is obtained. The same 1,000 pairs were therefore scored in three further ways on the same weights. The first is a prompted logit: the probability of ``yes'' over ``no'' when the model, without the adapter, is asked whether the function contains a security vulnerability, without generation. The second is a released activation oracle \citep{karvonen2025oracles}, an adapter that answers questions about activations injected at its second layer, given the activations of the function's last fifty tokens at its middle layer and the same question (Appendix~\ref{app:methods}). The third is a linear probe on per-layer mean-pooled activations, fitted with project-disjoint folds, with tf-idf over code tokens on the same folds as baseline. All four methods were scored as within-pair ranking, i.e. whether the vulnerable side scores above its own patch (Appendix~\ref{app:methods}). Ranking is easier than the absolute verdict the benchmarks require: it needs a direction that moves with the patch and no threshold, and absolute accuracy is not reported.

\begin{figure}[t]
\centering
\includegraphics{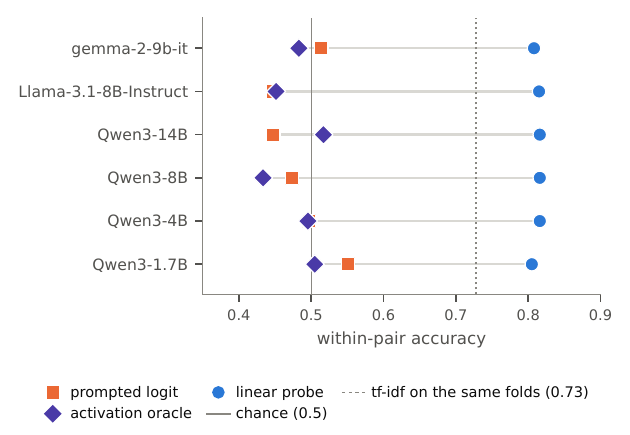}
\caption{The probe separates the pairs that the prompted logit and the activation oracle do not. Within-pair accuracy (vulnerable side scored above its patch) of the three methods on the same 1,000 pairs for the six models with an activation oracle.}
\label{fig:readouts}
\end{figure}

Scored on one scale, within-pair ranking with ties as coin flips, the generated verdict is at 0.50 to 0.52 for Llama-3.1-8B, gemma-2-9b and Qwen3-1.7B and at 0.55 to 0.57 for Qwen3-4B, 8B and 14B, the prompted logit at 0.45 to 0.55 for all six, and the probe at 0.81 to 0.82 (Figure~\ref{fig:readouts}; Table~\ref{tab:readouts}, Appendix~\ref{sec:length}); one of the six, gemma-2-9b-it, has an 8,192-token context and a 4,096-token budget. The activation oracle is at 0.43 to 0.52; it answers ``no'' for 60 to 95\% of inputs in three of the six models and favours the patched side on the bounds-check question by only $+0.004$ to $+0.051$ (Table~\ref{tab:channels}), so it barely registers the added check. A probe on the same activations reaches 0.81 to 0.82. Across 67 checkpoints with activations, the 61 population models and six further checkpoints (two base models, gemma-4-31B and its abliterated variant, a gemma-3-12b variant and OLMo-2-7B), probe accuracy at the best layer lies between 0.758 and 0.834. tf-idf on the same folds gives 0.728 and shuffled labels 0.42 to 0.50, and a rule that labels the shorter side vulnerable gives 0.91, above every probe, since the patched side is the longer one in 87.5\% of pairs; the comparison that means anything is therefore on length-matched pairs, below. Probe accuracy is nearly equal across checkpoints whose verdicts differ.

\paragraph{Length.} The probe could be reading length. Two checks were made on the probe trained on all pairs (Appendix~\ref{sec:length}). On 564 pairs whose sides differ in length by under 5\%, the probe is at 0.72 to 0.83 across 67 models, with a median of 0.78, and tf-idf at 0.64. It is tf-idf whose within-pair score follows the length ratio: its correlation with the log length ratio is $-0.59$, and the correlation for the probe is $-0.33$. On 125 pairs where the patch removed code, the vulnerable side is longer. There the probe falls to 0.41 to 0.64, with a median of 0.55, and tf-idf to 0.45. The probe therefore depends on code that a patch adds, and within that scope uses content that length does not explain.

\paragraph{Scope of the probe result.} A direction that separates the sides of a pair is linearly decodable from the activations for patches that add code, and the model's own outputs, generated or logit, do not move with it. Whether the model uses that direction was not tested; that would need an intervention along it, as has been done for truthfulness directions \citep{li2023iti,marks2024geometry}, and probe accuracy without an intervention is descriptive \citep{elazar2021amnesic}.

\section{Discussion}
\label{sec:disc}

Three statements follow. First, verdicts are determined by the text common to both sides. For 37 of 68 models the net score is within its interval of the null's zero while both-flagged and both-cleared rates exceed the null's, and for 64 of 68 both sides of a pair receive the same answer beyond independence. Whether a model flags or clears nearly everything depends on its family and on how its verdict is obtained. Second, the flag rate removes most of the difference between models of very different size: its contribution $p(1-p)$ does not depend on size, the input-dependent net score lies between 0.03 and 0.22 for every model that has one, and the best 27B open model equals the best API model. Third, the low scores are not explained by the absence of usable information in the function text. Some pairs cannot be decided from the function alone \citep{risse2025topscore}, and no human accuracy on these pairs was measured. A linear probe separates 0.78 of length-matched pairs by ranking, and 0.76 to 0.83 of all pairs, from activations of models whose generated verdicts rank at 0.50 to 0.57. The direction it finds follows code that the patch adds, a regularity of the corpus \citep{risse2024limits} as much as a property of the model. The result therefore shows a label-correlated signal that is linearly decodable and not reflected in the verdict.

These statements bear on benchmark design. A function-level or commit-level score measures flag rate more than model, and a pair-level score removes the flag rate's contribution and leaves a small number with wide intervals. A representative score would require pairs selected for decidability from the function text. Two questions remain open for interpretability work: whether an intervention along the probe direction changes the generated verdict, and what the dependence on shared text corresponds to in the weights. For reporting: pair-correct with all four outcome rates, the net score, the null of Equation~\ref{eq:null} and a bootstrap interval; no function-level score on its own; ranking of two models only where a paired interval of their difference excludes zero; a required verdict line with a deterministic parse together with the disagreement rate of a judge against it, since for a model whose text contradicts its verdict line the parse scores the line and not the model.

\section{Limitations}

Pair-correct is low everywhere, so results say little about which model is best. Sampling noise was estimated by drawing one of the three samples per side at random. It is a standard deviation of 0.009 to 0.022 in pair-correct at $n$=200 and 0.004 to 0.011 at $n$=1,000. Differences of 0.03 between models on a 200-pair benchmark are therefore at the noise level; the estimate is for one sample per side and bounds the noise of the majority of three from above. The probe is supervised, and the direction it learns corresponds to code that a patch adds: on the 125 pairs whose patch removes code it is at chance. The probe, the logit and the oracle are scored by within-pair ranking, and absolute accuracy is not reported. The probe was trained on all pairs and evaluated on the length-matched subsets; a probe trained on length-matched pairs only, and a probe on a randomly initialised network of the same architecture \citep{wieting2019random}, are not included. All pairs are C/C++ functions. One prompt was used, and prompt wording is a known source of variance in this literature that was not measured; temperature 0.7 and $K$=3 are choices of the same kind as the three varied here and were held fixed. Four of the five released benchmarks and ARVO are public and predate every model evaluated, so contamination cannot be excluded for any of them.

\begin{ack}
This work used the Helios supercomputer at ACK Cyfronet AGH within a Proof of Concept AI task of the LUMI AI Factory Service Centre (HORIZON-JU-EUROHPC-2025-AI-01-IBA).
\end{ack}

\bibliographystyle{plainnat}
\bibliography{refs}

\appendix

\section{Prompt}
\label{app:prompt}
System: \emph{You are a security vulnerability analyst. Given a function, decide whether it contains a memory-safety or security vulnerability. Begin your answer with a line that is exactly `Verdict: vulnerable' or `Verdict: not vulnerable', then briefly justify: the bug class, the violated invariant, and the code path, or why the function is safe.} User: \emph{Analyze this function for a security vulnerability.} followed by the function text. The parse takes the first verdict line after any visible reasoning block.

\section{Benchmarks}
\label{app:benches}
\begin{table}[h]
\centering
\caption{Pair construction and counts. Pairs: pairs in the benchmark after exact-duplicate removal. Evaluated: pairs scored per model; for \corpus{}, 1,000 for the panel and all 5,705 for the two full-corpus runs.}
\label{tab:benches}
\small
\begin{tabular}{llrr}
\toprule
Benchmark & Source of pairs & Pairs & Evaluated \\
\midrule
PrimeVul paired test split & security-fix commit, before/after function & 435 & 200 \\
VulnPatchPairs & security-fix commit, before/after function & 1,146 & 200 \\
PairVul & kernel security-fix commit, test split & 578 & 200 \\
JitVul & CVE-linked commit pairs & 808 & 200 \\
SecLLMHolmes & hand-curated real and synthetic pairs & 33 & 33 \\
Pooled set & PrimeVul, ARVO, CVE-fixing commits; two tiers & 5,705 & 1,000; 5,705 \\
\bottomrule
\end{tabular}
\end{table}

\section{The input-independent null per model}
\label{app:null}
\begin{figure}[h]
\centering
\includegraphics{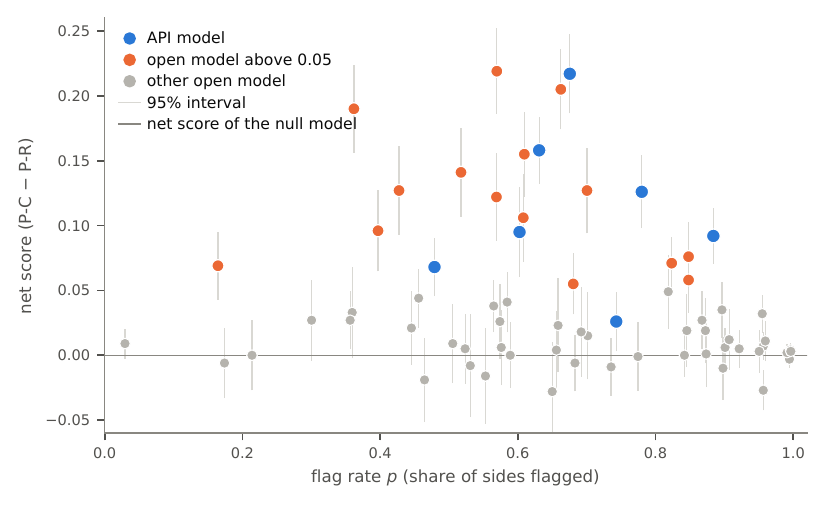}
\caption{Most models are within noise of the null model at every flag rate. Every model on the 1,000-pair draw of \corpus{} by flag rate $p$ and net score. Vertical bars are 95\% intervals of the net score.}
\label{fig:population}
\end{figure}
Table~\ref{tab:null} lists every model of the paper against the null model of Equation~\ref{eq:null} that flags each side with the model's own rate $p$: the observed both-flagged and both-cleared rates next to $p^2$ and $(1-p)^2$, the input-driven part P-C minus P-R, and Cohen's $\kappa$ between the two verdicts of a pair. No model has a both-flagged or both-cleared rate below the null's value, and for 37 of the 68 the 95\% interval of P-C minus P-R includes zero; the excess of the both-flagged and both-cleared rates over the null is the agreement between the two sides that patching does not remove. The four $\kappa$ values in italics belong to three models that flag every side and to one distillation at $p = 0.91$.

\begin{table}[h]
\centering
\caption{Every model against the null model that flags each side with the model's own rate $p$ regardless of the side: predicted both-flagged $p^2$ and both-cleared $(1-p)^2$ next to the observed rates, and the net score, P-C minus P-R; all rates are over parsed pairs. $\kappa$: agreement between the two verdicts of a pair beyond chance, with each side's own flag rate; values in italics are not distinguishable from zero (Fisher's exact test, $p \geq 0.05$). Sorted by P-C minus P-R.}
\label{tab:null}
\resizebox{0.92\textwidth}{!}{\begin{tabular}{lccccccc}
\toprule
Model & flag rate $p$ & net score & P-V & $p^2$ & P-B & $(1-p)^2$ & Cohen's $\kappa$ \\
\midrule
Qwen3.6-27B-abliterated & 0.57 & +0.219 & 0.404 & 0.324 & 0.265 & 0.185 & 0.36 \\
grok-4.3 (API) & 0.68 & +0.217 & 0.530 & 0.456 & 0.179 & 0.105 & 0.37 \\
Qwen3.6-27B & 0.66 & +0.205 & 0.519 & 0.440 & 0.192 & 0.113 & 0.38 \\
Qwen3.6-35B-A3B-abliterated-v4 & 0.36 & +0.191 & 0.196 & 0.132 & 0.469 & 0.405 & 0.30 \\
opus-4.8 (API) & 0.63 & +0.158 & 0.533 & 0.398 & 0.271 & 0.136 & 0.59 \\
Qwen3.6-35B-A3B & 0.61 & +0.155 & 0.460 & 0.373 & 0.238 & 0.152 & 0.38 \\
Qwen3-14B & 0.52 & +0.141 & 0.355 & 0.268 & 0.319 & 0.232 & 0.36 \\
deepseek-r1 (API) & 0.79 & +0.127 & 0.677 & 0.621 & 0.101 & 0.045 & 0.35 \\
qwen3-8b-heretic & 0.43 & +0.127 & 0.267 & 0.183 & 0.411 & 0.327 & 0.35 \\
gpt-oss-20b & 0.70 & +0.127 & 0.551 & 0.491 & 0.150 & 0.090 & 0.30 \\
Qwen3-8B & 0.57 & +0.122 & 0.413 & 0.324 & 0.275 & 0.186 & 0.37 \\
Qwen3-30B-A3B & 0.61 & +0.106 & 0.451 & 0.370 & 0.234 & 0.153 & 0.35 \\
Qwen3-4B & 0.40 & +0.096 & 0.264 & 0.158 & 0.470 & 0.364 & 0.45 \\
qwen3-8b (API) & 0.60 & +0.095 & 0.443 & 0.364 & 0.237 & 0.158 & 0.34 \\
gpt-5.2 (API) & 0.88 & +0.092 & 0.819 & 0.781 & 0.051 & 0.013 & 0.38 \\
GLM-4.7-Flash & 0.85 & +0.077 & 0.760 & 0.731 & 0.050 & 0.021 & 0.25 \\
gemma-4-26B-A4B-it & 0.82 & +0.071 & 0.767 & 0.678 & 0.120 & 0.031 & 0.61 \\
DeepSeek-R1-0528-Qwen3-8B & 0.86 & +0.069 & 0.766 & 0.738 & 0.048 & 0.020 & 0.24 \\
qwen3-4b-heretic & 0.16 & +0.069 & 0.072 & 0.027 & 0.743 & 0.698 & 0.33 \\
qwen3-coder (API) & 0.48 & +0.068 & 0.411 & 0.229 & 0.453 & 0.271 & 0.73 \\
NVIDIA-Nemotron-3-Nano-30B-A3B-BF16 & 0.85 & +0.058 & 0.766 & 0.723 & 0.065 & 0.022 & 0.34 \\
Olmo-3-7B-Think & 0.82 & +0.049 & 0.717 & 0.679 & 0.068 & 0.031 & 0.26 \\
VulnLLM-R-7B & 0.68 & +0.046 & 0.613 & 0.460 & 0.257 & 0.104 & 0.70 \\
gemma-4-E4B-it & 0.46 & +0.044 & 0.388 & 0.209 & 0.473 & 0.294 & 0.72 \\
phi-4 & 0.60 & +0.042 & 0.522 & 0.354 & 0.332 & 0.164 & 0.70 \\
Phi-4-mini-reasoning & 0.92 & +0.036 & 0.856 & 0.842 & 0.020 & 0.007 & 0.18 \\
Qwen3-1.7B & 0.36 & +0.033 & 0.198 & 0.131 & 0.473 & 0.407 & 0.29 \\
OpenVul-Qwen3-4B-GRPO & 0.96 & +0.032 & 0.929 & 0.916 & 0.015 & 0.002 & 0.32 \\
gemma-4-E4B-it-uncensored-heretic & 0.36 & +0.027 & 0.297 & 0.131 & 0.572 & 0.407 & 0.72 \\
Qwen2.5-Coder-7B-Instruct & 0.30 & +0.027 & 0.173 & 0.091 & 0.570 & 0.488 & 0.39 \\
Hermes-4.3-36B & 0.87 & +0.027 & 0.804 & 0.754 & 0.067 & 0.017 & 0.44 \\
deepseek-v3.2 (API) & 0.74 & +0.026 & 0.674 & 0.552 & 0.188 & 0.066 & 0.64 \\
granite-4.1-8b & 0.57 & +0.026 & 0.467 & 0.329 & 0.319 & 0.181 & 0.56 \\
NVIDIA-Nemotron-3-Nano-4B-BF16 & 0.66 & +0.023 & 0.487 & 0.434 & 0.169 & 0.116 & 0.24 \\
Phi-4-mini-instruct & 0.45 & +0.021 & 0.343 & 0.201 & 0.447 & 0.304 & 0.58 \\
Foundation-Sec-8B-Reasoning & 0.89 & +0.019 & 0.808 & 0.796 & 0.024 & 0.012 & 0.12 \\
DeepSeek-R1-Distill-Llama-8B & 0.85 & +0.019 & 0.746 & 0.719 & 0.050 & 0.023 & 0.21 \\
Ministral-8B-Instruct-2410 & 0.69 & +0.018 & 0.538 & 0.483 & 0.149 & 0.093 & 0.26 \\
Dolphin3.0-Llama3.1-8B & 0.70 & +0.015 & 0.554 & 0.492 & 0.151 & 0.089 & 0.30 \\
DeepSeek-R1-Distill-Qwen-7B & 0.92 & +0.012 & 0.844 & 0.839 & 0.012 & 0.007 & \emph{0.07} \\
SmolLM3-3B & 0.96 & +0.011 & 0.933 & 0.930 & 0.004 & 0.001 & 0.08 \\
Ministral-3-8B-Reasoning-2512 & 0.55 & +0.010 & 0.420 & 0.304 & 0.317 & 0.201 & 0.47 \\
OpenHermes-2.5-Mistral-7B & 0.03 & +0.009 & 0.012 & 0.001 & 0.953 & 0.942 & 0.39 \\
gemma-4-12B-it & 0.96 & +0.007 & 0.942 & 0.915 & 0.029 & 0.002 & 0.65 \\
granite-3.3-8b-instruct & 0.58 & +0.006 & 0.467 & 0.332 & 0.315 & 0.180 & 0.55 \\
gemma-3-4b-it & 0.90 & +0.006 & 0.871 & 0.812 & 0.069 & 0.010 & 0.66 \\
Ministral-3-3B-Instruct-2512 & 0.96 & +0.005 & 0.929 & 0.918 & 0.014 & 0.002 & 0.29 \\
Mistral-7B-Instruct-v0.3 & 0.53 & +0.005 & 0.435 & 0.282 & 0.372 & 0.220 & 0.61 \\
gpt-oss-20b-BF16-abliterated & 1.00 & +0.004 & 0.992 & 0.992 & 0.000 & 0.000 & \emph{-0.00} \\
DeepSeek-Coder-V2-Lite-Instruct & 0.66 & +0.004 & 0.537 & 0.430 & 0.225 & 0.118 & 0.47 \\
Llama-3.1-8B-Instruct & 0.95 & +0.003 & 0.918 & 0.909 & 0.011 & 0.002 & 0.20 \\
DeepHat-V1-7B & 1.00 & +0.003 & 0.993 & 0.993 & 0.000 & 0.000 & \emph{-0.00} \\
Ministral-3-14B-Instruct-2512 & 0.99 & +0.002 & 0.986 & 0.984 & 0.002 & 0.000 & 0.24 \\
Llama-3.2-3B-Instruct & 0.87 & +0.001 & 0.788 & 0.763 & 0.041 & 0.016 & 0.23 \\
Hermes-3-Llama-3.1-8B & 0.21 & +0.000 & 0.119 & 0.046 & 0.691 & 0.617 & 0.44 \\
gemma-3-12b-it & 0.84 & +0.000 & 0.808 & 0.710 & 0.122 & 0.025 & 0.74 \\
Ministral-3-8B-Instruct-2512 & 0.69 & +0.000 & 0.589 & 0.470 & 0.218 & 0.099 & 0.55 \\
Yi-Coder-9B-Chat & 0.78 & -0.001 & 0.691 & 0.613 & 0.124 & 0.047 & 0.46 \\
Qwen2.5-Coder-1.5B-Instruct & 0.99 & -0.003 & 0.989 & 0.989 & 0.000 & 0.000 & \emph{-0.01} \\
Foundation-Sec-1.1-8B-Instruct & 0.17 & -0.006 & 0.079 & 0.030 & 0.731 & 0.682 & 0.34 \\
gemma-2-9b-it & 0.69 & -0.006 & 0.627 & 0.475 & 0.248 & 0.097 & 0.71 \\
Qwen2.5-Coder-3B-Instruct & 0.53 & -0.008 & 0.325 & 0.283 & 0.262 & 0.219 & 0.17 \\
Olmo-3-7B-Instruct & 0.76 & -0.009 & 0.697 & 0.582 & 0.171 & 0.056 & 0.64 \\
Llama-3.2-3B-Instruct-heretic & 0.90 & -0.010 & 0.822 & 0.806 & 0.026 & 0.010 & 0.17 \\
Qwen3-8B-Base & 0.57 & -0.017 & 0.389 & 0.330 & 0.241 & 0.181 & 0.24 \\
Foundation-Sec-8B-Instruct & 0.46 & -0.019 & 0.326 & 0.216 & 0.396 & 0.286 & 0.44 \\
Meta-Llama-3.1-8B-Instruct-abliterated & 0.96 & -0.027 & 0.925 & 0.915 & 0.012 & 0.002 & 0.25 \\
WhiteRabbitNeo-2.5-Qwen-2.5-Coder-7B & 0.65 & -0.028 & 0.463 & 0.423 & 0.163 & 0.122 & 0.18 \\
\bottomrule
\end{tabular}
}
\end{table}

\section{Pair-correct minus reversed across the panel}
\label{app:net}
Table~\ref{tab:net} gives the net score of Section~\ref{sec:metric} for every benchmark and model. The reversed rate runs from 0.00 to 0.12; qwen3-8b has the highest or joint highest reversed rate on every benchmark, which places it below opus-4.8 and grok-4.3 on the net score where it ranks above them on pair-correct on four benchmarks. One value is negative (deepseek-v3.2 on PrimeVul, $-0.03$, with a reversed rate of 0.10), and the largest value is grok-4.3 on \corpus{}, 0.22.

\begin{table}[h]
\centering
\caption{Pair-correct minus reversed per benchmark and model, with the reversed rate in parentheses; same outputs as Table~\ref{tab:panel}.}
\label{tab:net}
\resizebox{\textwidth}{!}{\begin{tabular}{lccccccc}
\toprule
Benchmark & opus-4.8 & gpt-5.2 & grok-4.3 & deepseek-r1 & deepseek-v3.2 & qwen3-coder & qwen3-8b \\
\midrule
PrimeVul & 0.05 (0.04) & 0.07 (0.01) & 0.14 (0.04) & 0.14 (0.04) & -0.03 (0.10) & 0.05 (0.01) & 0.06 (0.11) \\
VulnPatchPairs & 0.10 (0.03) & 0.02 (0.06) & 0.09 (0.07) & 0.02 (0.10) & 0.01 (0.08) & 0.05 (0.07) & 0.04 (0.10) \\
PairVul & 0.15 (0.04) & 0.05 (0.07) & 0.11 (0.07) & 0.08 (0.10) & 0.07 (0.09) & 0.06 (0.01) & 0.02 (0.12) \\
JitVul & 0.08 (0.04) & 0.07 (0.01) & 0.13 (0.04) & 0.10 (0.05) & 0.07 (0.04) & 0.04 (0.05) & 0.17 (0.06) \\
SecLLMHolmes & 0.15 (0.00) & 0.03 (0.03) & 0.00 (0.09) & 0.09 (0.03) & 0.09 (0.03) & 0.03 (0.06) & 0.15 (0.09) \\
Ours & 0.16 (0.02) & 0.09 (0.02) & 0.22 (0.04) & 0.13 (0.05) & 0.03 (0.06) & 0.07 (0.03) & 0.09 (0.11) \\
\bottomrule
\end{tabular}
}
\end{table}

\section{Function-level F1 across the panel}
\label{app:f1}
F1 lies between 0.36 and 0.68 for every combination of benchmark and model, with a spread of 0.11 to 0.23 across the seven models within a benchmark. gpt-5.2 has the highest F1 on five of the six benchmarks and opus-4.8 on SecLLMHolmes; qwen3-coder has the lowest on every benchmark. On no benchmark is the model with the highest F1 the model with the highest pair-correct (Table~\ref{tab:panel}).

\begin{table}[h]
\centering
\caption{Function-level F1 on the same outputs as Table~\ref{tab:panel}.}
\resizebox{\textwidth}{!}{
\begin{tabular}{lccccccc}
\toprule
Benchmark & opus-4.8 & gpt-5.2 & grok-4.3 & deepseek-r1 & deepseek-v3.2 & qwen3-coder & qwen3-8b \\
\midrule
PrimeVul & 0.60 & 0.66 & 0.61 & 0.63 & 0.53 & 0.43 & 0.53 \\
VulnPatchPairs & 0.54 & 0.60 & 0.56 & 0.56 & 0.52 & 0.41 & 0.55 \\
PairVul & 0.56 & 0.56 & 0.55 & 0.56 & 0.55 & 0.36 & 0.48 \\
JitVul & 0.61 & 0.67 & 0.63 & 0.64 & 0.61 & 0.47 & 0.59 \\
SecLLMHolmes & 0.68 & 0.65 & 0.60 & 0.65 & 0.63 & 0.57 & 0.61 \\
Ours & 0.63 & 0.67 & 0.67 & 0.66 & 0.61 & 0.52 & 0.59 \\
\bottomrule
\end{tabular}
}
\end{table}

\section{The open-model population}
\label{app:pop}
\begin{figure}[h]
\centering
\includegraphics[width=0.92\textwidth]{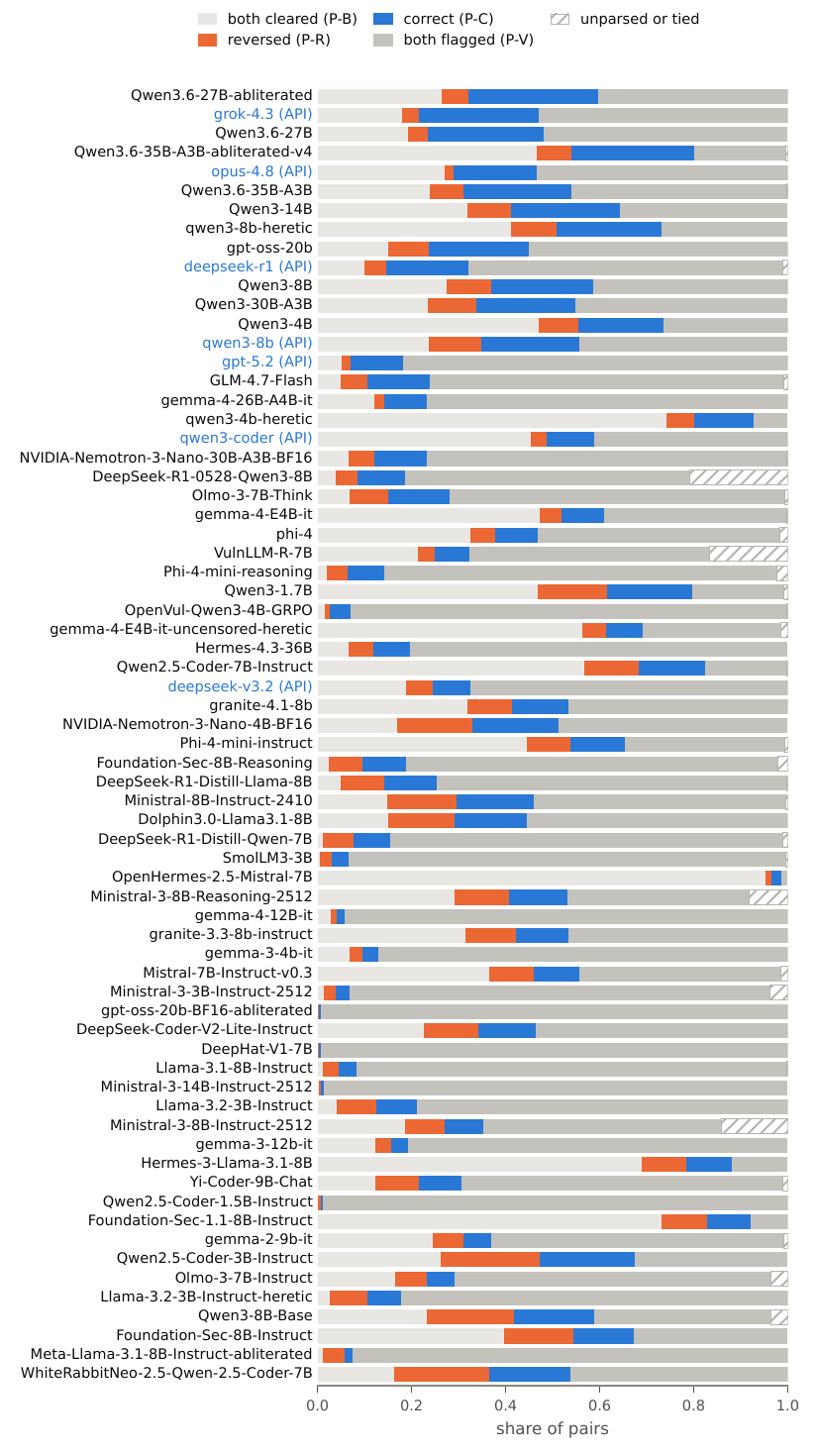}
\caption{The four outcomes of every model on the 1,000-pair draw of \corpus{}, sorted by pair-correct minus reversed; API models in blue; the hatched remainder is pairs unparsed or tied.}
\label{fig:cells}
\end{figure}

Figure~\ref{fig:cells} shows the four outcomes of every model sorted by pair-correct minus reversed, Table~\ref{tab:population} the roster and Table~\ref{tab:popcells} the outcome rates and F1. Fifty models have the protocol budget of 32,768 output tokens, 9 have 28,672 (a 32,768-token context), phi-4 has 12,288 and gemma-2-9b-it 4,096. 22 of the 61 models close a reasoning block in at least half of their samples. The unparsed-sample rate is below 5\% for all but four models: VulnLLM-R-7B (27\%), the R1-0528 distill (19\%), Ministral-3-8B (11\%) and Ministral-3-3B (5\%). The output text of the R1-0528 distill contained byte-level tokenizer markers in place of spaces and newlines; the markers were replaced by the characters they stand for before parsing, and the raw output is retained.

\begin{table}[h]
\centering
\caption{The self-hosted models. Reasoning block: share of samples containing a completed reasoning block; gpt-oss-20b returns its reasoning separately from the sample text. Budget: the effective output budget in tokens. No verdict line: share of samples without one.}
\label{tab:population}
\resizebox{\textwidth}{!}{
\begin{tabular}{llllrr}
\toprule
Model & family & group & reasoning block & budget (tokens) & no verdict line \\
\midrule
Qwen3-8B-Base & qwen3 & base & 0.00 & 28672 & 0.020 \\
DeepSeek-Coder-V2-Lite-Instruct & deepseek & code & 0.00 & 32768 & 0.000 \\
Qwen2.5-Coder-1.5B-Instruct & qwen2.5 & code & 0.00 & 28672 & 0.000 \\
Qwen2.5-Coder-3B-Instruct & qwen2.5 & code & 0.00 & 28672 & 0.000 \\
Qwen2.5-Coder-7B-Instruct & qwen2.5 & code & 0.00 & 28672 & 0.002 \\
Yi-Coder-9B-Chat & yi & code & 0.00 & 32768 & 0.006 \\
Hermes-4.3-36B & hermes & community & 0.00 & 32768 & 0.002 \\
Hermes-3-Llama-3.1-8B & llama & community & 0.00 & 32768 & 0.000 \\
Dolphin3.0-Llama3.1-8B & llama & community & 0.00 & 32768 & 0.000 \\
OpenHermes-2.5-Mistral-7B & mistral & community & 0.00 & 28672 & 0.001 \\
gemma-2-9b-it & gemma & general & 0.00 & 4096 & 0.008 \\
gemma-3-12b-it & gemma & general & 0.00 & 32768 & 0.000 \\
gemma-3-4b-it & gemma & general & 0.00 & 32768 & 0.000 \\
gemma-4-12B-it & gemma & general & 0.00 & 32768 & 0.000 \\
gemma-4-26B-A4B-it & gemma & general & 0.00 & 32768 & 0.000 \\
gemma-4-E4B-it & gemma & general & 0.00 & 32768 & 0.003 \\
GLM-4.7-Flash & glm & general & 0.89 & 32768 & 0.005 \\
gpt-oss-20b & gpt-oss & general & 0.00 & 32768 & 0.000 \\
granite-3.3-8b-instruct & granite & general & 0.00 & 32768 & 0.000 \\
granite-4.1-8b & granite & general & 0.00 & 32768 & 0.000 \\
Llama-3.1-8B-Instruct & llama & general & 0.00 & 32768 & 0.001 \\
Llama-3.2-3B-Instruct & llama & general & 0.00 & 32768 & 0.000 \\
Ministral-3-14B-Instruct-2512 & ministral & general & 0.00 & 32768 & 0.004 \\
Ministral-3-3B-Instruct-2512 & ministral & general & 0.00 & 32768 & 0.051 \\
Ministral-3-8B-Instruct-2512 & ministral & general & 0.00 & 32768 & 0.111 \\
Ministral-3-8B-Reasoning-2512 & ministral & general & 0.00 & 32768 & 0.043 \\
Ministral-8B-Instruct-2410 & ministral & general & 0.00 & 32768 & 0.001 \\
Mistral-7B-Instruct-v0.3 & mistral & general & 0.00 & 28672 & 0.012 \\
NVIDIA-Nemotron-3-Nano-30B-A3B-BF16 & nemotron & general & 1.00 & 32768 & 0.002 \\
NVIDIA-Nemotron-3-Nano-4B-BF16 & nemotron & general & 0.62 & 32768 & 0.001 \\
Olmo-3-7B-Instruct & olmo & general & 0.00 & 32768 & 0.028 \\
Olmo-3-7B-Think & olmo & general & 1.00 & 32768 & 0.007 \\
Phi-4-mini-instruct & phi & general & 0.00 & 32768 & 0.002 \\
Phi-4-mini-reasoning & phi & general & 0.98 & 32768 & 0.028 \\
phi-4 & phi & general & 0.00 & 12288 & 0.011 \\
Qwen3-1.7B & qwen3 & general & 1.00 & 32768 & 0.002 \\
Qwen3-14B & qwen3 & general & 1.00 & 32768 & 0.001 \\
Qwen3-30B-A3B & qwen3 & general & 1.00 & 32768 & 0.000 \\
Qwen3-4B & qwen3 & general & 1.00 & 32768 & 0.000 \\
Qwen3-8B & qwen3 & general & 1.00 & 32768 & 0.001 \\
Qwen3.6-27B & qwen3.6 & general & 1.00 & 32768 & 0.000 \\
Qwen3.6-35B-A3B & qwen3.6 & general & 1.00 & 32768 & 0.001 \\
SmolLM3-3B & smollm & general & 0.99 & 32768 & 0.005 \\
gemma-4-E4B-it-uncensored-heretic & gemma & abliterated & 0.00 & 32768 & 0.009 \\
gpt-oss-20b-BF16-abliterated & gpt-oss & abliterated & 0.00 & 32768 & 0.000 \\
Meta-Llama-3.1-8B-Instruct-abliterated & llama & abliterated & 0.00 & 32768 & 0.000 \\
Llama-3.2-3B-Instruct-heretic & llama & abliterated & 0.00 & 32768 & 0.000 \\
qwen3-4b-heretic & qwen3 & abliterated & 1.00 & 32768 & 0.000 \\
qwen3-8b-heretic & qwen3 & abliterated & 1.00 & 32768 & 0.000 \\
Qwen3.6-35B-A3B-abliterated-v4 & qwen3.6 & abliterated & 1.00 & 32768 & 0.003 \\
Qwen3.6-27B-abliterated & qwen3.6 & abliterated & 1.00 & 32768 & 0.000 \\
DeepSeek-R1-Distill-Qwen-7B & deepseek & reasoning-distilled & 1.00 & 32768 & 0.006 \\
DeepSeek-R1-Distill-Llama-8B & llama & reasoning-distilled & 1.00 & 32768 & 0.001 \\
DeepSeek-R1-0528-Qwen3-8B & qwen3 & reasoning-distilled & 1.00 & 32768 & 0.187 \\
DeepHat-V1-7B & deephat & security & 0.00 & 28672 & 0.000 \\
Foundation-Sec-1.1-8B-Instruct & foundation-sec & security & 0.00 & 32768 & 0.000 \\
Foundation-Sec-8B-Instruct & foundation-sec & security & 0.00 & 32768 & 0.000 \\
Foundation-Sec-8B-Reasoning & foundation-sec & security & 0.79 & 32768 & 0.017 \\
OpenVul-Qwen3-4B-GRPO & qwen3 & security & 1.00 & 32768 & 0.001 \\
VulnLLM-R-7B & vulnllm & security & 0.43 & 28672 & 0.267 \\
WhiteRabbitNeo-2.5-Qwen-2.5-Coder-7B & whiterabbitneo & security & 0.00 & 28672 & 0.000 \\
\bottomrule
\end{tabular}
}
\end{table}
\begin{table}[h]
\centering
\caption{Population outcome rates and function-level F1 on the 1,000-pair draw, protocol parse; columns as in Table~\ref{tab:cells}.}
\label{tab:popcells}
\resizebox{\textwidth}{!}{
\begin{tabular}{llcccccc}
\toprule
Model & group & P-C & P-V & P-B & P-R & unparsed or tied & function-level F1 \\
\midrule
Qwen3-8B-Base & base & 0.170 & 0.375 & 0.232 & 0.186 & 0.037 & 0.526 \\
DeepSeek-Coder-V2-Lite-Instruct & code & 0.121 & 0.537 & 0.225 & 0.117 & 0.000 & 0.569 \\
Qwen2.5-Coder-1.5B-Instruct & code & 0.004 & 0.989 & 0.000 & 0.007 & 0.000 & 0.664 \\
Qwen2.5-Coder-3B-Instruct & code & 0.202 & 0.325 & 0.262 & 0.210 & 0.001 & 0.512 \\
Qwen2.5-Coder-7B-Instruct & code & 0.142 & 0.172 & 0.568 & 0.115 & 0.003 & 0.393 \\
Yi-Coder-9B-Chat & code & 0.091 & 0.683 & 0.123 & 0.092 & 0.011 & 0.610 \\
Hermes-4.3-36B & community & 0.078 & 0.803 & 0.067 & 0.051 & 0.001 & 0.645 \\
Hermes-3-Llama-3.1-8B & community & 0.095 & 0.119 & 0.690 & 0.095 & 0.001 & 0.300 \\
Dolphin3.0-Llama3.1-8B & community & 0.155 & 0.554 & 0.151 & 0.140 & 0.000 & 0.590 \\
OpenHermes-2.5-Mistral-7B & community & 0.022 & 0.012 & 0.952 & 0.013 & 0.001 & 0.064 \\
gemma-2-9b-it & general & 0.059 & 0.621 & 0.246 & 0.065 & 0.009 & 0.578 \\
gemma-3-12b-it & general & 0.035 & 0.807 & 0.122 & 0.035 & 0.001 & 0.628 \\
gemma-3-4b-it & general & 0.033 & 0.871 & 0.069 & 0.027 & 0.000 & 0.645 \\
gemma-4-12B-it & general & 0.018 & 0.942 & 0.029 & 0.011 & 0.000 & 0.659 \\
gemma-4-26B-A4B-it & general & 0.092 & 0.767 & 0.120 & 0.021 & 0.000 & 0.649 \\
gemma-4-E4B-it & general & 0.091 & 0.387 & 0.472 & 0.047 & 0.003 & 0.501 \\
GLM-4.7-Flash & general & 0.132 & 0.754 & 0.050 & 0.056 & 0.008 & 0.661 \\
gpt-oss-20b & general & 0.213 & 0.551 & 0.150 & 0.086 & 0.000 & 0.636 \\
granite-3.3-8b-instruct & general & 0.112 & 0.467 & 0.315 & 0.106 & 0.000 & 0.538 \\
granite-4.1-8b & general & 0.120 & 0.467 & 0.319 & 0.094 & 0.000 & 0.547 \\
Llama-3.1-8B-Instruct & general & 0.037 & 0.915 & 0.011 & 0.034 & 0.003 & 0.657 \\
Llama-3.2-3B-Instruct & general & 0.086 & 0.788 & 0.041 & 0.085 & 0.000 & 0.636 \\
Ministral-3-14B-Instruct-2512 & general & 0.007 & 0.985 & 0.002 & 0.005 & 0.001 & 0.666 \\
Ministral-3-3B-Instruct-2512 & general & 0.030 & 0.894 & 0.013 & 0.025 & 0.038 & 0.658 \\
Ministral-3-8B-Instruct-2512 & general & 0.083 & 0.506 & 0.187 & 0.083 & 0.141 & 0.576 \\
Ministral-3-8B-Reasoning-2512 & general & 0.125 & 0.385 & 0.291 & 0.116 & 0.083 & 0.525 \\
Ministral-8B-Instruct-2410 & general & 0.165 & 0.536 & 0.148 & 0.147 & 0.004 & 0.589 \\
Mistral-7B-Instruct-v0.3 & general & 0.098 & 0.428 & 0.366 & 0.093 & 0.015 & 0.519 \\
NVIDIA-Nemotron-3-Nano-30B-A3B-BF16 & general & 0.113 & 0.764 & 0.065 & 0.055 & 0.003 & 0.651 \\
NVIDIA-Nemotron-3-Nano-4B-BF16 & general & 0.183 & 0.487 & 0.169 & 0.160 & 0.001 & 0.579 \\
Olmo-3-7B-Instruct & general & 0.059 & 0.672 & 0.165 & 0.068 & 0.036 & 0.598 \\
Olmo-3-7B-Think & general & 0.131 & 0.712 & 0.068 & 0.082 & 0.007 & 0.641 \\
Phi-4-mini-instruct & general & 0.115 & 0.341 & 0.444 & 0.094 & 0.006 & 0.484 \\
Phi-4-mini-reasoning & general & 0.078 & 0.836 & 0.020 & 0.043 & 0.023 & 0.660 \\
phi-4 & general & 0.092 & 0.513 & 0.326 & 0.051 & 0.018 & 0.562 \\
Qwen3-1.7B & general & 0.180 & 0.196 & 0.469 & 0.147 & 0.008 & 0.441 \\
Qwen3-14B & general & 0.233 & 0.355 & 0.319 & 0.092 & 0.001 & 0.578 \\
Qwen3-30B-A3B & general & 0.210 & 0.451 & 0.234 & 0.104 & 0.001 & 0.597 \\
Qwen3-4B & general & 0.181 & 0.264 & 0.470 & 0.085 & 0.000 & 0.496 \\
Qwen3-8B & general & 0.217 & 0.413 & 0.275 & 0.095 & 0.000 & 0.589 \\
Qwen3.6-27B & general & 0.247 & 0.518 & 0.192 & 0.042 & 0.001 & 0.658 \\
Qwen3.6-35B-A3B & general & 0.228 & 0.459 & 0.238 & 0.073 & 0.002 & 0.619 \\
SmolLM3-3B & general & 0.037 & 0.928 & 0.004 & 0.026 & 0.005 & 0.662 \\
gemma-4-E4B-it-uncensored-heretic & abliterated & 0.078 & 0.292 & 0.563 & 0.051 & 0.016 & 0.437 \\
gpt-oss-20b-BF16-abliterated & abliterated & 0.006 & 0.992 & 0.000 & 0.002 & 0.000 & 0.667 \\
Meta-Llama-3.1-8B-Instruct-abliterated & abliterated & 0.018 & 0.925 & 0.012 & 0.045 & 0.000 & 0.647 \\
Llama-3.2-3B-Instruct-heretic & abliterated & 0.071 & 0.822 & 0.026 & 0.081 & 0.000 & 0.639 \\
qwen3-4b-heretic & abliterated & 0.127 & 0.072 & 0.742 & 0.058 & 0.001 & 0.300 \\
qwen3-8b-heretic & abliterated & 0.224 & 0.267 & 0.411 & 0.097 & 0.001 & 0.530 \\
Qwen3.6-35B-A3B-abliterated-v4 & abliterated & 0.262 & 0.195 & 0.467 & 0.072 & 0.004 & 0.530 \\
Qwen3.6-27B-abliterated & abliterated & 0.275 & 0.404 & 0.265 & 0.056 & 0.000 & 0.635 \\
DeepSeek-R1-Distill-Qwen-7B & reasoning-distilled & 0.077 & 0.836 & 0.012 & 0.065 & 0.010 & 0.651 \\
DeepSeek-R1-Distill-Llama-8B & reasoning-distilled & 0.111 & 0.744 & 0.050 & 0.092 & 0.003 & 0.636 \\
DeepSeek-R1-0528-Qwen3-8B & reasoning-distilled & 0.101 & 0.607 & 0.038 & 0.046 & 0.208 & 0.658 \\
DeepHat-V1-7B & security & 0.005 & 0.993 & 0.000 & 0.002 & 0.000 & 0.667 \\
Foundation-Sec-1.1-8B-Instruct & security & 0.092 & 0.079 & 0.731 & 0.098 & 0.000 & 0.254 \\
Foundation-Sec-8B-Instruct & security & 0.129 & 0.326 & 0.396 & 0.148 & 0.001 & 0.473 \\
Foundation-Sec-8B-Reasoning & security & 0.092 & 0.790 & 0.023 & 0.073 & 0.022 & 0.649 \\
OpenVul-Qwen3-4B-GRPO & security & 0.044 & 0.927 & 0.015 & 0.012 & 0.002 & 0.668 \\
VulnLLM-R-7B & security & 0.073 & 0.511 & 0.214 & 0.035 & 0.167 & 0.593 \\
WhiteRabbitNeo-2.5-Qwen-2.5-Coder-7B & security & 0.173 & 0.463 & 0.163 & 0.201 & 0.000 & 0.553 \\
\bottomrule
\end{tabular}
}
\end{table}

\section{Pair-level results in the literature}
\label{app:lit}
The published tables show the same gap wherever both numbers are printed. PrimeVul's fine-tuned 7B models reach F1 18 to 21 with P-C 1.6 to 3.0 and P-B 84 to 88 (their Table V). In Vul-RAG's table the uniform ``all vulnerable'' guess has the best F1 (0.67) with pair accuracy 0.00, and LineVul has 0.64 against 0.02 \citep{du2024vulrag}. On JitVul GPT-4o's plain prompt gives F1 65.96 with pair accuracy 1.02 \citep{jitvul2025}. On OpenVul's 985 pairs F1 is about twice the pair rate for every model \citep{openvul2026}.

The pair-level literature shows the same. On PrimeVul's pairs GPT-4 with chain of thought scored P-C 12.94 in 2024 (Table VIII of \citealp{ding2024primevul}); the panel here on the same benchmark is at 0.07 to 0.18, with the 8B model at 0.17. On JitVul the best configurations give GPT-4o-mini 20.2, GPT-4o 19.1 and Llama-3.1-8B 15.2 \citep{jitvul2025}. On PairVul, GPT-4o, Claude 3.5 Sonnet, Qwen2.5-Coder-32B and DeepSeek-V2 lie within six points of one another \citep{du2024vulrag}. On CWE-Trace every model from 4B to GPT-4.1-mini averages 49 to 54\% over the two halves of its pairs, because each answers one way for both \citep{zibaeirad2026directional}. A frontier gap of 10 to 15 points appears where the pair score was judged on the explanation (OpenVul's judge-scored P-Pass@1: Qwen3-4B 7.5, DeepSeek-V3.1 22.5 \citep{openvul2026}; the reasoning-verified counts of \citealp{lu2026reasoning}), and a GRPO-tuned 4B reaches 21.1.

The order among frontier models changes from paper to paper, and the best untuned result on any released pair set is about 22\% (PrimeVul GPT-4 CoT 12.9; OpenVul's best zero-shot 22.5; VulTrial's four-agent GPT-4o 18.6 on 435 PrimeVul pairs, \citealp{widyasari2025vultrial}). The one result far above these, 64\% on the same 435 PrimeVul pairs, comes from a multi-agent pipeline that first synthesises a behavioural contract for the code, and no single model given a function reaches it \citep{wang2026phoenix}.

\section{The abliterated variants}
\label{app:twins}
The abliterated variants keep their base's discrimination within 0.03 in seven of eight cases and change its default answer in a direction that depends on the family: the Qwen variants toward ``safe'' (Qwen3-8B both-flagged 0.41 to 0.27 and both-cleared 0.28 to 0.41; Qwen3-4B 0.26 to 0.07 and 0.47 to 0.74; Qwen3.6-35B-A3B 0.46 to 0.20 and 0.24 to 0.47; Qwen3.6-27B 0.52 to 0.40 and 0.19 to 0.27), the gemma-4-E4B variant the same way (0.39 to 0.29), the two Llama variants by under 0.04, and the gpt-oss-20b variant to flagging everything (0.55 to 0.99) with its discrimination gone (0.13 to 0.00).

\section{Judge protocol}
\label{app:judges}
Every judge, API or self-hosted, received the same system prompt: \emph{You read a security analysis of one function. Reply with ONLY a JSON object \{``reached\_vulnerable'': true|false\}: does the analysis conclude the function contains an exploitable vulnerability. Judge only what the analysis concludes, not the code.} followed by the full analysis (the last 80,000 characters where longer) as the user turn; temperature 0; the answer is read from the JSON field. The API judges (grok-4.3, gpt-5-mini, claude-haiku-4.5) were given one sample per side of the seven API models' outputs on \corpus{}; the open judges all three samples per side of every model's pooled-set draw, served with vLLM, Llama-3.3-70B-Instruct in its FP8 release, gpt-oss-120b in its released MXFP4, Qwen3-32B in BF16 with thinking off. Agreement is measured on samples that contain a verdict line; pair-correct under judge extraction takes the majority of the judged samples per side. Parse failures: Ministral-3-8B writes its verdict line as \texttt{Verdict: **vulnerable**}, which the regular expression of the protocol, requiring the verdict word immediately after the colon, misses on 14.1\% of its sides; a variant that tolerates markup parses all but 0.5\% and changes its pair-correct from 0.083 to 0.114, and three other models by under a point. VulnLLM-R-7B writes its training format, \texttt{Correct Answer: vulnerable}, for 27\% of sides, the R1-0528 distill has no verdict line in 19\% of samples, for these a judge is the only extraction, and it gives 0.12 against 0.07 under the parse for VulnLLM-R and 0.16 to 0.20 against 0.10 for the distill. The API judges agreed with the parse on 99.2\% to 100\% of samples with a verdict line and changed pair-correct by $-0.016$ to $+0.034$ per model (the most for deepseek-r1, whose late verdicts the parse misses); where no verdict line existed they nevertheless answered, grok-4.3 ``not vulnerable'' on all 86 empty outputs. On the seven API models' analyses the open judges and the API judges agreed on 99.9 to 100\% of the samples with a verdict line (gpt-oss-120b: 96.8\%) and on 21 or 22 of the 22 without one.
\begin{table}[h]
\centering
\caption{The open judges. Agreement: median over models of the share of samples with a verdict line on which the judge agrees with the parse (range in parentheses). Vulnerable, not vulnerable, no answer: shares of the judge's answers on samples without a verdict line. Pair-correct, judge minus parse: median over models of pair-correct under judge extraction minus pair-correct under the parse (range).}
\label{tab:judges}
\resizebox{\textwidth}{!}{\begin{tabular}{lcccccc}
\toprule
Judge & agreement with the parse & vulnerable & not vulnerable & no answer & pair-correct, judge minus parse & models judged \\
\midrule
Llama-3.3-70B (FP8) & 0.999 (0.63 to 1.00) & 0.44 & 0.55 & 0.003 & +0.001 (-0.03 to +0.17) & 68 \\
gpt-oss-120b & 0.968 (0.66 to 1.00) & 0.57 & 0.43 & 0.002 & +0.001 (-0.11 to +0.15) & 68 \\
Qwen3-32B & 0.999 (0.63 to 1.00) & 0.45 & 0.54 & 0.002 & +0.001 (-0.00 to +0.17) & 68 \\
\bottomrule
\end{tabular}
}
\end{table}

\section{Length control for the probe}
\label{sec:length}

The patched side is the longer one in 87.5\% of pairs, so that a rule that labels the shorter side vulnerable reaches 0.875 without access to the code, and mean-pooled activations depend on length. Two checks were performed to separate length from content. On the 564 pairs whose two sides differ in length by under 5\%, the probe was at 0.72 to 0.83 across the 67 models (median 0.78) and tf-idf at 0.64; on the 319 pairs differing by 5 to 20\%, the probe was at 0.80 to 0.89 and tf-idf at 0.85, and on the 117 pairs differing by more the two were equal at approximately 0.85. The tf-idf baseline is the one that depends on the length difference (its within-pair score difference correlates with the log length ratio at $-0.59$, that of the probe at $-0.33$); on length-matched pairs the probe was 0.14 above it. The second check used the 125 pairs in which the vulnerable side is the longer one, i.e. in which the patch removed code: there the probe was at 0.41 to 0.64 (median 0.55) and tf-idf at 0.45, both near chance. The probe therefore depends on code that a patch adds and does not transfer to patches that remove code; within that scope, it uses content that is not explained by length.

\begin{table}[t]
\centering
\caption{Four ways of obtaining a verdict on the same 1,000 pairs. Verdict, net: the net score of the generated verdict under the protocol. Verdict, ranking: the same verdict as within-pair ranking with ties as coin flips, $0.5 + $ net$/2$. Logit, oracle, probe: within-pair ranking (vulnerable side scored above its patch) of the prompted logit, the activation oracle and the linear probe; chance is 0.5, tf-idf on the same folds 0.728.}
\label{tab:readouts}
\small
\begin{tabular}{lccccc}
\toprule
Model & verdict, net & verdict, ranking & logit & oracle & probe \\
\midrule
Qwen3-1.7B & +0.033 & 0.516 & 0.551 & 0.505 & 0.805 \\
Qwen3-4B & +0.096 & 0.548 & 0.497 & 0.495 & 0.816 \\
Qwen3-8B & +0.122 & 0.561 & 0.473 & 0.433 & 0.816 \\
Qwen3-14B & +0.141 & 0.571 & 0.448 & 0.517 & 0.816 \\
Llama-3.1-8B-Instruct & +0.003 & 0.501 & 0.448 & 0.452 & 0.815 \\
gemma-2-9b-it & -0.006 & 0.497 & 0.514 & 0.483 & 0.808 \\
\bottomrule
\end{tabular}

\end{table}

\section{Protocol for the four verdict methods}
\label{app:methods}
Activations for the probe: hidden states at every layer, mean-pooled over the function's tokens, input capped at 4,096 tokens. Probe: logistic regression on standardized features, $C$=0.5, five folds grouped by project, out-of-fold scores, within-pair ranking, 2,000 bootstrap resamples over pairs; baselines on the same folds: tf-idf over identifier and symbol tokens (50,000 features, sublinear tf) and function length. Oracle: the released adapters \texttt{adamkarvonen/checkpoints\_*} for Qwen3-1.7B, 4B, 8B, 14B, Llama-3.1-8B-Instruct and gemma-2-9b-it; activations of the last 50 tokens of each side at the adapter's middle layer (50\% depth), each side truncated after the last token that differs from its partner so that the 50 tokens include the patched region; added to the activations at layer 1 after scaling to the norm of the activations already there; questions: ``Does the code this activation comes from contain a security vulnerability? Answer Yes or No.'' and ``Does the code this activation comes from end with a bounds check or a null check? Answer Yes or No.''; the answer probability is the softmax over the ``Yes'' and ``No'' token groups at the answer position. Prompted logit: the base model with ``Does this code contain a security vulnerability? Answer Yes or No.'' after the function text, same scoring; the adapter-on-text control loads the oracle adapter without the injection hook.
\begin{table}[h]
\centering
\caption{Yes-rate on the vulnerable side when the base model is given the function text, when the oracle adapter is given the text, and when the adapter is given the activations; the last column is the difference in p(yes) on the bounds-check question, patched side minus vulnerable side.}
\label{tab:channels}
\small
\begin{tabular}{lcccc}
\toprule
Model & base, text & adapter, text & adapter, activations & check question \\
\midrule
Qwen3-1.7B & 1.00 & 0.99 & 1.00 & +0.023 \\
Qwen3-4B & 0.92 & 0.85 & 0.87 & +0.047 \\
Qwen3-8B & 0.98 & 0.89 & 0.32 & +0.051 \\
Qwen3-14B & 0.93 & 0.88 & 0.29 & +0.013 \\
Llama-3.1-8B-Instruct & 0.87 & 0.80 & 0.40 & +0.004 \\
gemma-2-9b-it & 0.90 & 0.25 & 0.05 & +0.046 \\
\bottomrule
\end{tabular}

\end{table}

\end{document}